\documentclass{article}
\pdfoutput=1
\usepackage[a4paper, total={6in, 9in}]{geometry}

\usepackage{main}
\usepackage[utf8]{inputenc}
\usepackage[T1]{fontenc}
\usepackage{lmodern}
\usepackage{hyperref}
\usepackage{url}
\usepackage{booktabs}
\usepackage{amsfonts}
\usepackage{nicefrac}
\usepackage{microtype}
\usepackage{color}
\usepackage[]{graphicx}
\usepackage{amsmath}
\usepackage{xcolor}
\usepackage{paralist}
\usepackage{enumerate}
\usepackage[]{graphicx}
\usepackage[]{subcaption}
\usepackage{threeparttable}
\usepackage[numbers]{natbib}

\usepackage{titling}

\usepackage{threeparttable}
\usepackage{tikz}
\usepackage{multirow}
            
\usepackage{array}

\usepackage{placeins}

\newcolumntype{C}[1]{>{\centering\let\newline\\\arraybackslash\hspace{0pt}}m{#1}}

\usepackage{caption} 
\newcommand{\arr}[2]{$#1 \rightarrow #2$}
\newcommand{\arrt}[2]{\texttt{#1}$\rightarrow$\texttt{#2}}

\newcommand{\draft}[1]{}

\pretitle{\begin{center} \LARGE \bf}
\posttitle{\par\end{center}\vskip 0.5em}
\preauthor{
    \begin{center}
    \normalsize \lineskip 0.5em%
    \begin{tabular}[t]{c}
}
\postauthor{\end{tabular}\par\end{center}}

\predate{\begin{center}\large}
\postdate{\par\end{center}}

\title{Generalization to Novel Objects Using Prior Relational Knowledge}

\author{
  Varun Kumar Vijay \\
  Intel AI Lab\\
  Santa Clara, CA \\
  \texttt{varun.v.kumar@intel.com}
  \and
  Abhinav Ganesh \\
  Intel AI Lab \\
  Santa Clara, CA \\
  \texttt{abhinav.ganesh@intel.com} \\
   \and
   Hanlin Tang \\
   Intel AI Lab \\
   Santa Clara, CA \\
   \texttt{hanlin.tang@intel.com} \\
   \and
   Arjun K. Bansal \\
   Intel AI Lab \\
   Santa Clara, CA \\
   \texttt{arjun.bansal@intel.com}
}

\date{}

\begin{document}
\maketitle

\begin{abstract}
To solve tasks in new environments involving objects unseen during training, agents must reason over prior information about those objects and their relations. We introduce the Prior Knowledge Graph network, an architecture for combining prior information, structured as a knowledge graph, with a symbolic parsing of the visual scene, and demonstrate that this approach is able to apply learned relations to novel objects whereas the baseline algorithms fail. Ablation experiments show that the agents ground the knowledge graph relations to semantically-relevant behaviors. In both a Sokoban game and the more complex Pacman environment, our network is also more sample efficient than the baselines, reaching the same performance in 5-10x fewer episodes. Once the agents are trained with our approach, we can manipulate agent behavior by modifying the knowledge graph in semantically meaningful ways. These results suggest that our network provides a framework for agents to reason over structured knowledge graphs while still leveraging gradient based learning approaches.
\end{abstract}

\section{Introduction}
\label{introduction}

Humans have a remarkable ability to both generalize known actions to novel objects, and reason about novel objects once their relationship to known objects is understood. For example, on being told a novel object (e.g. 'bees') is to be avoided, we readily apply our prior experience avoiding known objects without needing to experience a sting. Deep Reinforcement Learning (RL) has achieved many remarkable successes in recent years including results with Atari \cite{Mnih2015} games and Go \cite{Silver} that have matched or exceeded human performance. While a human playing Atari games can, with a few sentences of natural language instruction, quickly reach a decent level of performance, modern end-to-end deep reinforcement learning methods still require millions of frames of experience (for e.g. see Fig. 3 in \cite{Lake2016}). Past studies have hypothesized a role for prior knowledge in addressing this gap between human performance and Deep RL \cite{Dubey2018, Lake2016}. 

While other works have studied the problem of generalizing tasks involving the same objects (and relations) to novel environments, goals, or dynamics \cite{Finn2017, Nichol2018, Packer2018, Rusu2016, Wang2018, Zambaldi2018}, here we specifically study the problem of generalizing known relationships to novel objects. Zero-shot transfer of such relations could provide a powerful mechanism for learning to solve novel tasks. We speculated that objects might serve as a useful intermediate symbolic representation to combine the visual scene with knowledge graphs encoding the objects and their relations \cite{Janner2019}. 

To build this approach, we needed novel components to transfer information between the knowledge graph and the symbolic scene. Prior approaches \cite{yang2018visual, Ammanabrolu2018} have only used one-directional transfers, and without diverse edge relation types. In this paper, we propose the Prior Knowledge Graph Network (PKGNet), which makes several key contributions:

\begin{enumerate}
    \item We introduced new layer types (\texttt{Broadcast}, \texttt{Pooling} and \texttt{KG-Conv} for sharing representations between the knowledge graph and the symbolic visual scene. 
    \item We leveraged edge-conditioned convolution \cite{Simonovsky2017, Nassar18} to induce our method to learn edge-specific relations that can be applied to novel objects.
    \item Compared to several baselines (DQN \cite{Mnih2015}, DQN-Prioritized Replay \cite{schaul2015prioritized}, A2C \cite{mnih2016asynchronous}) in two environments (Sokoban, PacMan), our approach is 5-10x more sample efficient during training, and importantly, able to apply learned relations to novel objects.
    \item We describe a mechanistic role for how the knowledge graph is leveraged in solving the tasks.
\end{enumerate}

We observed agents' behavior while manipulating the knowledge graph during runtime (i.e., using a trained agent), which confirmed that those edges were grounded to the game semantics. These results demonstrate that our PKGNet framework can provide a foundation for faster learning and generalization in deep reinforcement learning.

\section{Related Work}

\subsection{Graph Based Reinforcement Learning}
Graph-based architectures for reinforcement learning have been applied in several contexts. In recent work in control problems and text-based games, the graph is used as a structured representation of the state space, either according to the anatomy of the agent \cite{Wang2018} or to build a description of relations in the world during text based exploration \cite{Ammanabrolu2018}. We use the graph as structured prior knowledge to inject into the network. 

In Yang et al \cite{yang2018visual}, a knowledge graph was used to assist agents in finding novel objects in a visual search task. Our work has several key differences. Here we test methods for applying learned relations (e.g. push, avoid, chase) to novel objects. In their approach, adding those explicit relations to the graph significantly impaired performance, so relations are omitted from their knowledge graph, rendering their approach not viable in our task. Yang et al. also rely on significantly more prior knowledge, including word embeddings and object co-occurrence. Instead, here we provide arbitrary one-hot relation vectors, and agents learn to relate those vectors to actions, while achieving better relative performance gains.

Previous approaches take a one-directional approach of concatenating the knowledge graph features into the state features. We hypothesize that reasoning in both feature space and structured representations is important, and therefore introduce components to share representations between the two domains. By pooling state features into the graph and performing convolutions, our model implements a global operation similar to the self-attention layer used in the Relational RL architecture \cite{Zambaldi2018}. However, that model tackles the problem of learning relational knowledge during training, without any \textit{a priori} knowledge. Our model is designed to exploit external knowledge to generalize to new objects at test time.

\subsection{Extracting symbols for Reinforcement Learning}
Several studies have extracted objects from visual input using unsupervised or semi-supervised methods \cite{Burgess2019, Chen2016, Cheung2014, Greff2019, Higgins2016, Higgins2017, Ionescu2018, Kingma2013, Locatello2018}.  
As the focus of our study is combining scene graphs and knowledge graphs, and not the extraction of symbols themselves, we assume that our network has object level ground truth information available from the scene. For this reason we use environments that can be programmatically generated. In comparison to approaches that operate in 3-D environments \cite{yang2018visual}, our approach solves fundamental problems of how and whether prior relational knowledge encoded with extremely minimal knowledge can be leveraged efficiently.
 
%\subsection{Natural language guided RL} 

%Previous work has used natural language instruction  to train agents \cite{Ammanabrolu2018,Coreyes2018,Fu2018,Kaplan2017}. Our approach is complementary to these as the information extraction algorithms in the literature (for e.g. \cite{Angeli2015}) could be used to structure natural language corpora, rules or instructions as knowledge graphs, which can be provided to PKG-DQN. We believe that graphs could more generally serve as an efficient and interpretable representational mechanism for prior knowledge or instruction.

%\subsection{Metalearning} 

%Several prior studies study generalization to novel tasks. Techniques such as MAML \cite{Finn2017} train on a set of environments or tasks and attempt to learn a weight manifold on which a new task can be learned with minimal additional examples. Here we structure the relevant information for solving the task in the form of a knowledge graph to avoid needing additional training. In environments where such prior knowledge may be unavailable, the PKG-DQN could be augmented with metalearning or curriculum learning. Other techniques such as DARLA \cite{Higgins2017} and progressive nets \cite{Rusu2016} have studied generalization in the context of domain shifts (e.g. from simulation to reality) but not in the context of generalization to new objects studied here.

\section{Prior Knowledge Graph Network}
\label{PKG-DQN}

\begin{figure}[ht]
\begin{center}
\centerline{\includegraphics[width=\linewidth]{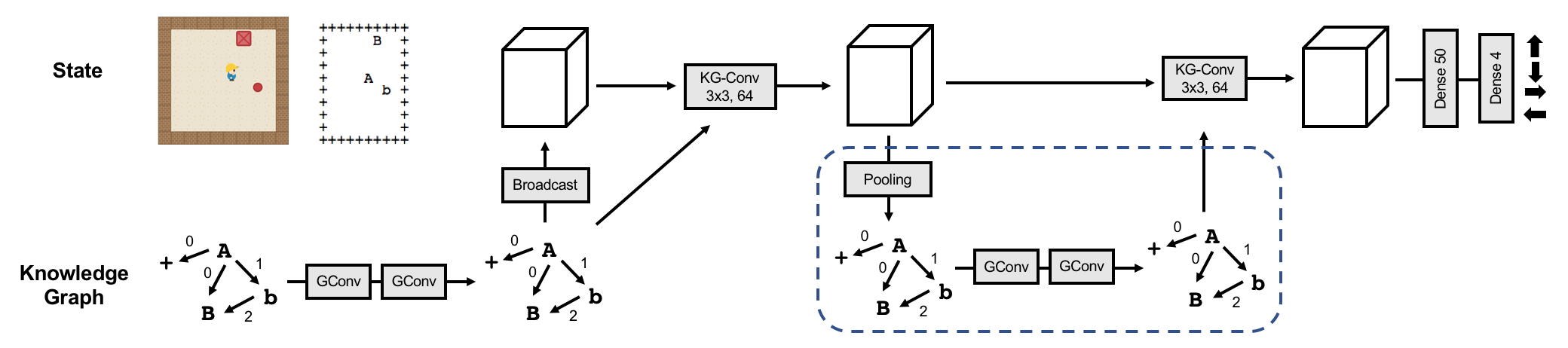}}
\caption{Prior Knowledge Graph network (PKGNet) architecture. The knowledge graph is first operated on by graph convolution layers (GConv) to enrich the node features to $d=64$ dimensions \cite{Simonovsky2017}. We then use \texttt{Broadcast} to create a compatible scene representation $S \in \mathbb{R}^{10\times10\times d}$, indicated here by the cube. The network trunk consists of several \texttt{KG-Conv} layers. The side branch (blue dotted region) allows for reasoning over the knowledge graph structure. See main text for a more detailed description.}
\label{fig:network}
\end{center}
\end{figure}

Reinforcement learning models often use RGB features as input to a Convolutional Neural Network (CNN). We apply our algorithms to symbolically parsed visual environments by encoding each symbol into a one-hot vector, as is done for character-level CNN models in natural language processing \cite{Zhang2015}. In our proposed PKGNet, the input consists of a knowledge graph (representing prior information) and a scene graph. 

\par{\bf State.} While PKGNet can handle general scene graphs, the environments in this paper are 2-D grid worlds, a specific subset of scene graphs where the vertices are the symbols, and the edges connect neighboring entities. Edge-conditioned graph convolution then reduces to regular 2D convolution. Therefore, we refer to the scene graph by its state representation $\mathcal{S} \in \mathbb{R}^{h \times w \times d}$, which is the feature map. 

\par{\bf Knowledge Graph.} The knowledge graph $\mathcal{K}=(\mathcal{V}, \mathcal{E})$ is a directed graph provided as vertices for each symbol in the environment (for subjects and objects), $\mathcal{V}=\{v_A, v_b, v_B, v_+, \ldots \}$ initially encoded as one-hot vectors of length $|\mathcal{V}|$, and edge features $\mathcal{E}=\{e_{Ab}, e_{AB}, e_{A+}, \ldots\}$. The edge features (for relations) are represented as one-hot vectors. The connectivity of the graph, as well as the edge features are designed to reflect the structure of the environment. During training, the knowledge graph's structure and features are fixed. Importantly, while we provide the one-hot encoded representation of the edge relationships, the agent must learn to ground the meaning of this representation in terms of rewarding actions during training. If successfully grounded, the agent may use this representation during the test phase when it encounters novel objects connected with known relationships to entities in the knowledge graph.

\par{\bf Algorithms.} We tested our network with the Deep-Q Network \cite{Mnih2015}, as well as Prioritized Experience Replay (PER) \cite{schaul2015prioritized}, and the A2C algorithm \cite{mnih2016asynchronous}.

\subsection{Model Architecture}
 The model architecture is shown in Figure \ref{fig:network}. First, we apply two layers of edge-conditioned graph convolution (ECC) \cite{Simonovsky2017} to $\mathcal{K}$ to enrich the node features with information from the neighborhood of each node. Those features are then encoded in the state representation $\mathcal{S}$ through a \texttt{Broadcast} layer. The network's main trunk consists of several \texttt{KG-Conv} layers, which serve to jointly convolve over the state and knowledge graph. The side branch (dotted blue rectangle), enables reasoning over the structured knowledge graph. In the side branch, we first update the knowledge graph with \texttt{Pooling} from the state, followed by graph convolutions. Then, we update the state representation with a \texttt{KG-Conv} layer, which incorporates the updated knowledge graph. Finally, for DQN and DQN-PER, we use a few linear layers to compute the Q-values for each action. For A2C, we also emit a value estimate. We provide below more details on the individual components.

\subsection{Model Components}
We introduce several operations for transferring information between the state representation $\mathcal{S}$ and the knowledge graph $\mathcal{K}$. We can \texttt{Broadcast} the knowledge graph features into the state, or use \texttt{Pooling} to gather the state features into the knowledge graph nodes. We can also update the state representation by jointly convolving over $(\mathcal{S}, \mathcal{K})$, which we call \texttt{KG-Conv}. Supplemental Figure 1 shows visual depictions of these operations.

\paragraph{\bf Graph Convolutions.} In order to compute features for the entities in the knowledge graph, we use an edge-conditioned graph convolution (ECC) \cite{Simonovsky2017}. In this formulation, a multilayer perceptron network is used to generate the filters given the edge features as input. Each graph layer $g$ computes the feature of each node $v_i$ as:

\begin{equation}
v_i = \sum_{v_j \in \mathcal{N}(v_i)} \Theta[e_{ij}]v_j + b
\end{equation}

\noindent where the weight function $\Theta[e_{ij}]$ depends only on the edge feature and is parameterized as a neural network. $\mathcal{N}(v_i)$ is the set of nodes with edges into $v_i$. Our implementation uses graph convolution layers with $d=64$ features, and the weight network is a single linear layer with 8 hidden units. The output of $g$ is a graph with nodes $\mathcal{V} \in \mathbb{R}^{|\mathcal{V}| \times d}$.

\paragraph{\bf Broadcast.} We define the function $\texttt{Broadcast}:\mathcal{K}\rightarrow\mathcal{S}$. For each entity $i$ in the knowledge graph, we copy its graph representation $v_i$ to each occurrence of $i$ in the game map. This is used to initialize the state representation $\mathcal{S}$ such that we are using a common embedding to refer to entities in both $\mathcal{K}$ and $\mathcal{S}$. 

% Each location $(i,j)$ in the state is computed as

% \begin{equation}
%     \mathcal{S}_{i,j} = \sum_{v \in \mathcal{V}} \delta_v(i,j) v
% \end{equation}

% \noindent where $\delta_v(i,j)=1$ if the entity corresponding to $v$ is present at location $(i,j)$ and zero otherwise. Thus, symbols in the game map not present in the knowledge graph are initialized with a zero vector.

\paragraph{\bf Pooling.} The reverse of \texttt{Broadcast}, this operation is used to update the entity representations in the knowledge graph. In $\texttt{Pooling}:\mathcal{S}\rightarrow\mathcal{K}$, we update the graph's representation $v$ by averaging the features in $\mathcal{S}$ over all instances of entity corresponding to $v$ in the state.

% \begin{equation}
%     v_i = \frac{1}{N_v}\sum_{(i,j) \in \mathcal{S}} W\delta_v(i,j)\mathcal{S}_{ij}
% \end{equation}

% \noindent where $N_v=\sum_\mathcal{S} \delta_v(i,j)$ is the number of instances of $v$ in the state. Since $\mathcal{S}$ and $\mathcal{V}$ may have different number of features, we used the weight matrix $W$ to project from the state vectors to the dimensionality of the vertex features in the graph.

\paragraph{\bf KG-Conv.} To update the state representation $\mathcal{S}$, we augment a regular convolution layer with the knowledge graph. In addition to applying convolutional filters to the neighborhood of a location, we also add the node representation $v_i$ of the entity $i$ at that location, passed through linear layer to $v_i$ to match the number of filters in the convolution. Formally, we can describe this operation as:

\begin{equation}
    \text{Conv}_{3\times 3 \times d}(\mathcal{S}) + \text{Conv}_{1\times 1 \times d}(\mathtt{Broadcast}(\mathcal{K}))
\end{equation}

This provides a skip connection allowing deeper layers in the network to more easily make use of the global representations.

\section{Experiments}
\label{experiments}

Previous environments measured generalization to more difficult levels \cite{cobbe2018quantifying, Nichol2018}, modified environment dynamics \cite{Packer2018}, or different solution paths \cite{Zambaldi2018}. These environments, however, do not introduce new objects at test time. To quantify the generalization ability of PKGNet to unseen objects, we needed a symbolic game with the ability to increment the difficulty in terms of the number of new objects and relationships. Therefore, we use a variation of the Sokoban environment, where the agent pushed balls into the corresponding bucket, and new ball and bucket objects and their pairing are provided at test time. We also benchmarked our model and the baseline algorithms on Pacman, after extracting a symbolic representation \cite{berkpac}.

\subsection{Sokoban}

A variant of the Sokoban environment is implemented using the \textit{pycolab} framework \cite{pycolab}. The set of rewarded ball-bucket pairs varies, and in the test games the agent sees balls or buckets not seen during training. For the variations, see Table \ref{tab:exp}. We increasingly vary the difficulty of the environment by the number of ball-bucket pairs, the complexity of the grouping, and the number of unseen objects. The buckets-repeat is a challenging environment, with complex relationships in the test environment.

\begin{table}[tb]
\caption{Experiment variations for the Sokoban environment. The agent is rewarded for pushing the ball into the correct bucket. For each type, we list the rewarded ball-bucket pairs in the training and test games. Note that the test games only include ball types not seen in the training games. Sets denote rewarded combinations. For example, $\{b, c\} \rightarrow B$ means \arr{b}{B} and \arr{c}{B} are rewarded.}\label{tab:exp}
\vskip 0.15in
\begin{center}
\begin{small}
\begin{tabular}{p{1.4cm}p{2.5cm}p{2.5cm}}
\toprule
Name & Training Pairs & Test Pairs \\
\midrule
one-one & $b\rightarrow B$ & $c\rightarrow B$ \\
two-one & $\{b, c\}\rightarrow B$ & $d\rightarrow B$ \\
five-two & $\{b, c, d, e, f\}\rightarrow B$ & $\{g, h\} \rightarrow B$  \\
\midrule
buckets & \arr{b}{B}, \arr{c}{C} , \arr{d}{D} , \arr{e}{E} , \arr{f}{F} & \arr{g}{G} , \arr{h}{H} , \arr{i}{I} , \arr{j}{J} , \arr{k}{K} \\
buckets-repeat & $\{b,c,d\}\rightarrow B$ , $\{e,f,g\}\rightarrow C$, \ldots ,  $\{n,o,p\}\rightarrow F$  & $\{q,r,s\}\rightarrow G$, $\{t,u,v\}\rightarrow H$ , \ldots , $\{6,7,8\}\rightarrow K$ \\
\midrule
\end{tabular}
\end{small}
\end{center}
\vskip -0.1in
\end{table}

\begin{figure}[tb]
\begin{center}
\centerline{\includegraphics[width=\linewidth]{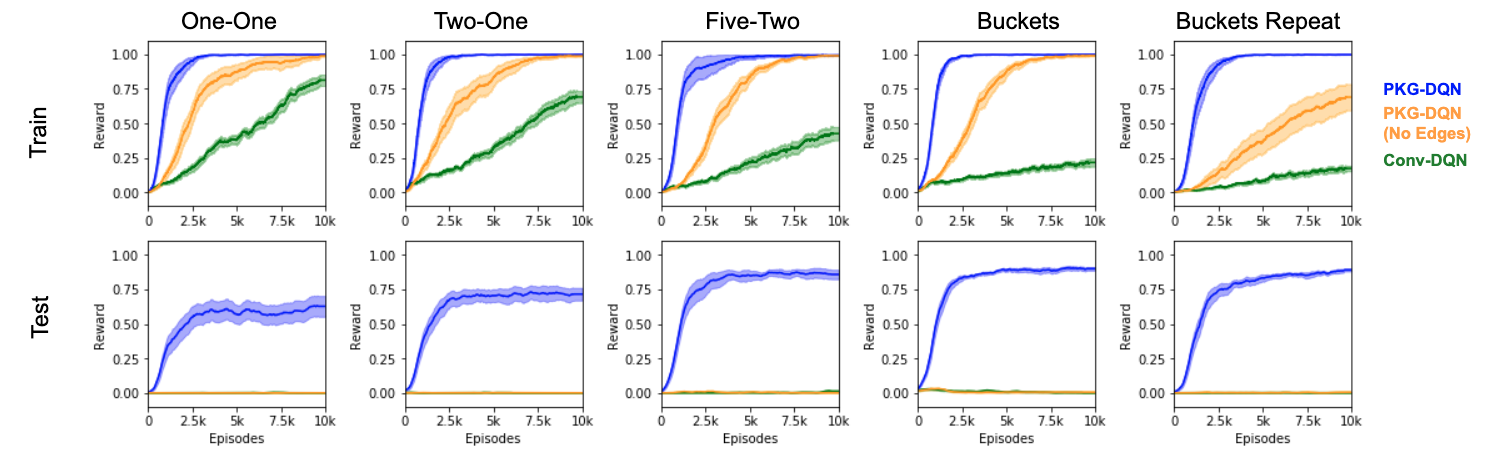}}
\caption{Sokoban results. For the environments described in Table \ref{tab:exp} (columns), performance of the baseline DQN (green), our proposed PKG-DQN (blue), and a variant of PKG-DQN with edges removed (orange) over the number of training episodes. Success rate (fraction of environments completed within 100 steps) is shown for training (top row) and test (bottom row) environments. Bold lines are the average over $n=10$ runs, and shaded area denotes the standard error. A moving average of $t=100$ episodes was applied.}
\label{fig:kg_perf}
\end{center}
\end{figure}

\begin{figure}[tb]
\begin{center}
\centerline{\includegraphics[width=\linewidth]{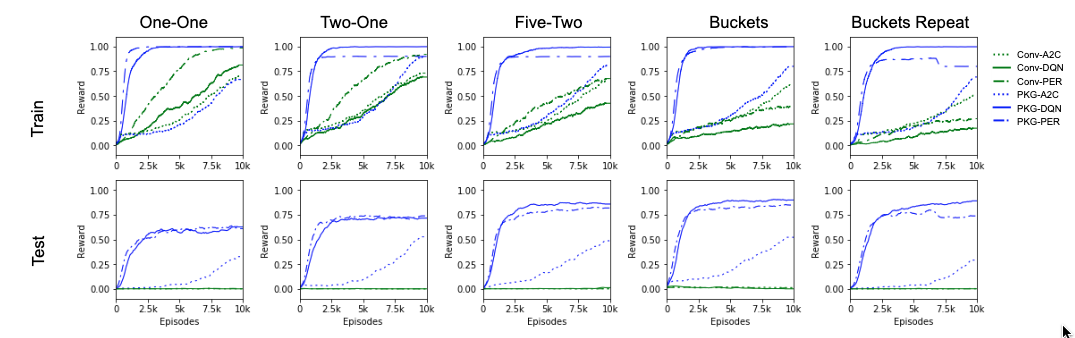}}
\caption{Algorithms. Our approach with the graph (blue) is more sample efficient in the training environment (first row) compared to the baselines (green) across all tested algorithms (DQN, DQN-PER, and A2C), as indicated by the line styles. The graph approach is also required for generalizing to novel objects in the test environment (second row).}
\label{fig:algos}
\end{center}
\end{figure}

\begin{figure}[tb]
\begin{center}
\centerline{\includegraphics[width=\linewidth]{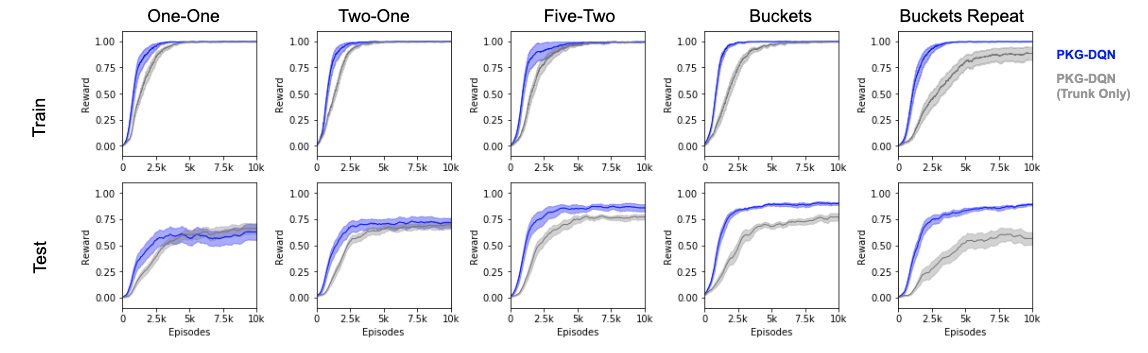}}
\caption{Model ablation. We compare the PKG-DQN model to a simplified version without the side branch. We note that performance, particularly on the test maps, drops significantly.}
\label{fig:modelab}
\end{center}
\end{figure}

\subsection{Pacman}
We test the agents on the smallGrid, mediumClassic, and capsuleClassic environments from Pacman. The environments differed in the size of the map as well as the numbers of ghosts, coins, and capsules present. The agent experienced +10 points for eating a coin, +200 for eating a ghost, +500 for finishing the coins, -500 for being eaten, and -1 for each move.

\subsection{Knowledge graph construction}
For both environments, we add all entities to the knowledge graph with the exception of blank spaces. We then add edges between objects to reflect relationships present in the game structure. Each entity or edge type is assigned a unique one-hot vector; note however that edges between two pairs of entities may have the same edge type if they are connected with a similar relationship. While we attach semantic meaning to these edge categories, their utility is grounded by the model during training. Additional details are in the supplement.

\section{Results}

Our network can be used in conjunction with a variety of RL algorithms. Here we tested PKG-DQN, PKG-A2C, and PKG-PER against the regular convolutional baselines in the Sokoban and Pacman environments. In addition, we compared the performance of different knowledge graph architectures during training. We also demonstrated the ability to manipulate agent behavior by changing the knowledge graph at test time.

\label{results}
\subsection{Sokoban}
In the Sokoban environment, the PKG-DQN model was more sample efficient during training than the baseline Conv-DQN algorithm, as shown in Figure \ref{fig:kg_perf}. For example, in the one-one environment, our model required approximately 8x fewer samples to reach the solution in the training environment. In addition, in more complex environments with an increased number of possible objects and ball-bucket pairings, the baseline Conv-DQN required increasingly more samples to solve, whereas the PKG-DQN solved in the same number of samples. 

We tested zero-shot transfer learning by placing the trained agents in environments with objects unseen during training. The PKG-DQN is able to leverage the knowledge graph to generalize, solving in $> 80\%$ of the test environments (see Figure \ref{fig:kg_perf}, bottom row). The baseline DQN failed completely to generalize to these environments.

When we deleted the edges from the PKG-DQN (Figure \ref{fig:kg_perf}, orange lines), the model trained slower and failed to generalize. The No Edge condition still trained faster than the baseline Conv-DQN, possibly due to additional parameters in the \texttt{KG-Conv}, however, that advantage is minimal in our most complex Sokoban environment, the buckets-repeat. We also tested baselines with significantly more parameters and different learning rates without improvement.

These observations held across all baselines tested (DQN, DQN - Prioritized Replay, and A2C), as shown in Figure \ref{fig:algos}. The relative performance of the Conv baselines is consistent with previous results (for e.g. Ms. Pacman in Table S3 in \cite{mnih2016asynchronous}). We also ran an ablation study where we removed the side branch from Figure \ref{fig:network} (blue dotted rectangle), which significantly impacted sample efficiency and generalization (Figure \ref{fig:modelab}). This demonstrates that the structured reasoning, enabled by the bi-directional flow of information from our novel layers, is important for performance.

\subsection{Knowledge graph types}

In order to determine whether the results in Figure \ref{fig:kg_perf} are sensitive to the choice of the knowledge graph architecture, we trained the PKG-DQN model with variants of the base knowledge graph, as shown in Figure \ref{fig:kg_types}: `Base' (graph cropped to entities present in the scene), graphs with same (`Same Edges') or no (`No Edges') edges, a fully connected graph either the same edge label (`Fully Connected') or distinct edge labels (`Fully Connected - Distinct'), and a `Complete' graph with no cropping based on presence in the scene.

When we removed the edge distinctiveness (`Same Edges'), the model still trained, but failed to generalize to novel objects. If we removed edges entirely (`No Edges'), the performance is the same as the baseline DQN. These results show that encoding the game structure into the knowledge graph is important for generalizing to the test environment but not necessary for the training environments.

Surprisingly, when the knowledge graph is fully connected (`FC' and `FC-distinct'), the model does not train, suggesting that the prior structure cannot be learned by PKG-DQN. If the complete graph is available during training, including nodes for objects that only appear in the test environments, the model generalizes to near-optimal performance (see orange lines in `Complete'). In this condition, even though the object `c' is not in the training environment, gradients still flow through the \arr{A}{c} edge. To avoid any contamination during training into the knowledge graph of information about ball-bucket object pairs seen during test, for the base condition we crop the knowledge graph only to entities (and corresponding edges) seen in the training environments.

\subsection{Pacman}
\begin{figure}[!tb]
\begin{center}
\centerline{\includegraphics[width=\linewidth]{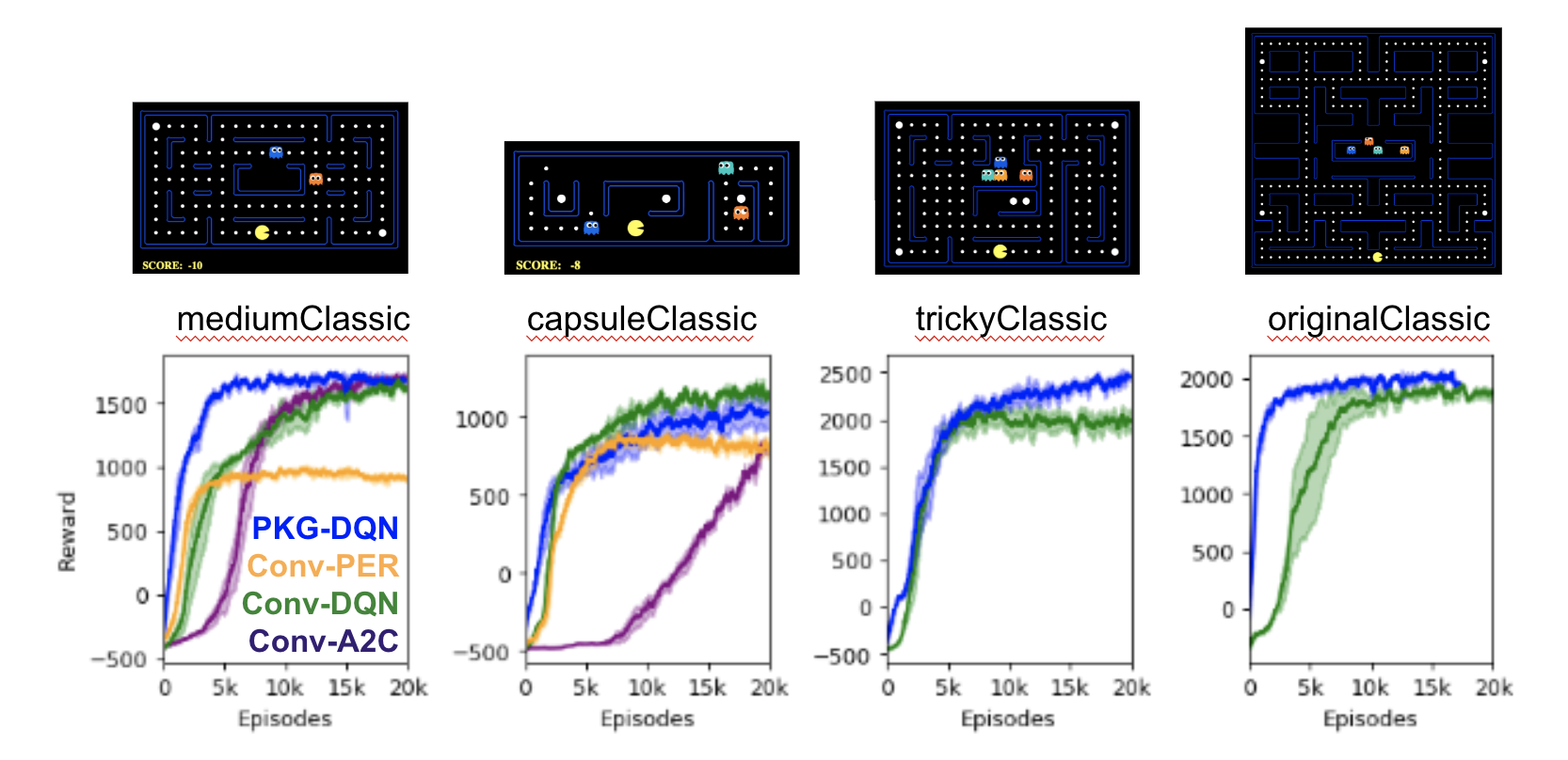}}
\caption{Pacman results. Performance of the baseline Conv-based models (green, purple, orange) and our PKG-DQN (blue) agent on several Pacman environments (smallGrid, mediumClassic, and capsuleClassic). Bold lines are the mean.}
\label{fig:pacman}
\end{center}
\end{figure}

The PKG-DQN converges significantly faster to a performing control policy than the convolution-based DQN on all three Pacman environments (Figure \ref{fig:pacman}). Both models reach similar levels of final performance on smaller environments, which is expected, as the convolutional model should eventually be able to deduce the relations between the symbols with enough training.

\subsection{What do the agents learn?}

\begin{figure}[tb]
\begin{center}
\centerline{\includegraphics[width=0.9\linewidth]{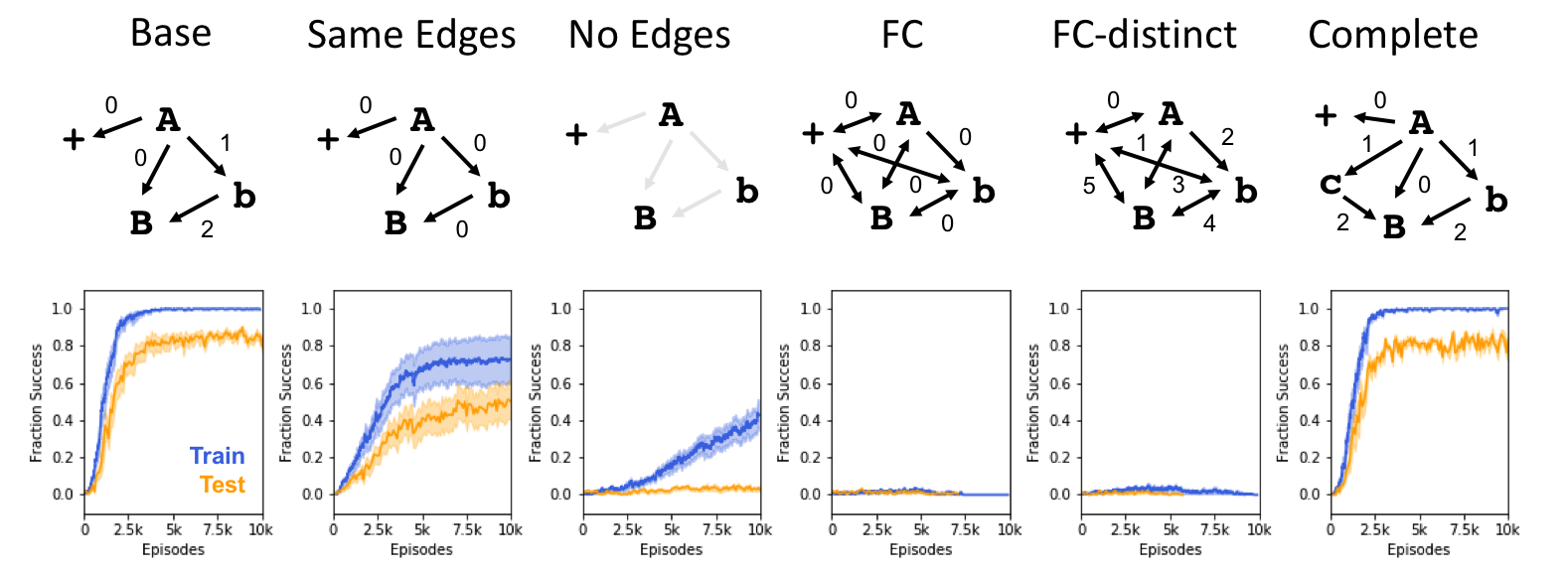}}
\caption{Performance of the PKG-DQN when trained with several knowledge graph variants in our most challenging environment (buckets-repeat). Train performance in blue, and test performance in orange. Shaded errors indicate the standard error over $n=10$ repetitions. Results in other environments are similar, but omitted for space reasons. See Supplement for additional environments.}
\label{fig:kg_types}
\end{center}
\end{figure}

\begin{figure}[!tb]
\begin{center}
\centerline{\includegraphics[width=0.9\linewidth]{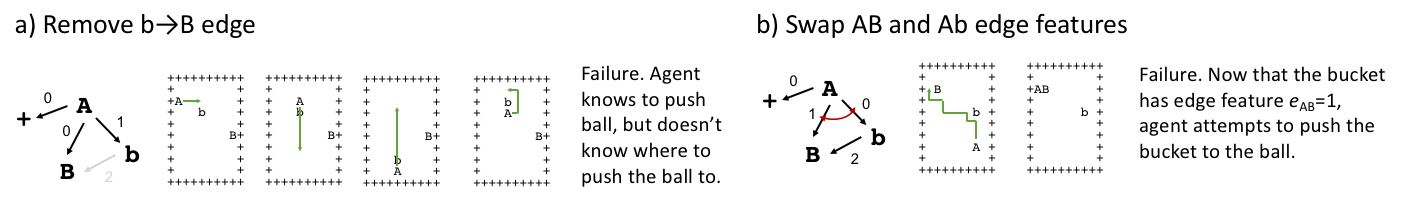}}
\caption{Manipulating Trained Agents in Sokoban. We used agents trained on the base knowledge graph. and manipulated their behavior at runtime by changing the input knowledge graph.}
\label{fig:manip}
\end{center}
\end{figure}

To understand how the agents are interpreting the edge relations between objects, we observed the behavior of a trained agent running in an environment while manipulating the knowledge graph (Figure \ref{fig:manip}). For simplicity consider the one-one environment, with one bucket pair (\arr{b}{B}) during training and one pair (\arr{c}{B}) during testing. When we removed \arr{b}{B}, the agent still pushes the ball, but does not know where to push the ball towards, suggesting that the agent has learned to ground the feature $e_{bB}=2$ as 'goal' or 'fills'. We swapped the edge features of \arr{A}{B} and \arr{A}{b}, and the agent attempts to push the bucket into the ball. The knowledge graph could also be manipulated such the agent pushes a ball into another ball (Supplement). These studies show that the agent learned the 'push' and 'fills' relation and can apply these actions to objects it has never pushed before.

Similarly, in Pacman, if we remove the \arrt{Player}{Scared Ghost} edge, the agent no longer chases the scared ghosts (Table \ref{tab:manip_pacman}). Without an edge to the capsule, the agent no longer eats the capsule. The agent can also be manipulated to not avoid ghosts by changing the \arrt{Ghost}{Player} feature to the \arrt{Player}{Coin} edge relation.

\begin{table*}[tb]
    \centering
    \begin{small}
    \caption{Manipulating Pacman behavior. Behavior and score of the PKG-DQN agent on the mediumClassic map when various edges are removed or features substituted. Reward is shown as mean$\pm$ standard error over $n=100$ repetitions.}     \label{tab:manip_pacman}

    \vskip 0.15in

    \begin{tabular}{|l|r|l|}
    \hline
    Variation & Reward & Behavior \\
    \hline
    Base & $1169\pm39$ & Default behavior\\
    Set \arrt{Ghost}{Player} to \arrt{Player}{Coin} feature & $-78\pm38$ & Does not avoid ghosts\\
    Remove \arrt{Player}{Scared Ghost} edge & $558\pm25$ & Does not chase scared ghosts\\
    Remove \arrt{Player}{Coin} edge & $-376\pm20$ & Pacman moves randomly\\
    Remove \arrt{Player}{Capsule} edge & $323\pm37$ & Does not eat the capsule\\
    Remove \arrt{Player}{Wall} edge & $-339\pm21$ & Runs into the nearest wall\\
    Remove \arrt{Scared Ghost}{Wall} edge & $530\pm28$ & Does not chase scared ghosts\\
    \hline
    \end{tabular}
    \end{small}
\end{table*}

\section{Discussion}
\label{discussion}

We've demonstrated the efficacy of a general approach to augmenting networks with knowledge graphs that facilitate faster learning, and more critically enable algorithms to apply their learned relations to novel objects. This was substantiated with experiments across two environments and multiple algorithms. Ablation studies highlight the importance of bi-directional information exchange between the state features and knowledge graph, in contrast to previous work with one-directional feature concatenation \cite{yang2018visual, Ammanabrolu2018}. Moreover, the use of edge-conditioned convolution allows the agent to ground and leverage edge relations, as well as generalize to changing knowledge graphs. Previous approaches \cite{yang2018visual} are most similar to our `Same Edges` case (Figure \ref{fig:kg_types}), which was significantly worse performing.

% Compared to the Relational Deep RL model \cite{Zambaldi2018} which is a general approach that learns from all pairwise relationships in the scene, but scales as $\mathit{\mathcal{O}(number\_of\_nodes^{2})}$, our approach assumes spatial relations only between nearby locations in the scene graph, and connects entities with semantic relationships in the knowledge graph. Through hierarchy we pool information along the edges of the scene and knowledge graphs. Our approach could therefore scale as $\mathit{\mathcal{O}(number\_of\_nodes * number\_of\_edges)}$.

Our approach is complementary to other approaches in RL that strive to improve sample efficiency and generalization such as hierarchical RL \cite{Kulkarni2016}, metalearning \cite{Finn2017}, or better exploration policies \cite{Clune2018} and can be combined as such with these approaches to build better overall systems. Interestingly, attempts to learn the knowledge graph during training were not successful (see 'fully connected' in Figure \ref{fig:kg_types}), and we speculate that graph attention models \cite{Velickovic2018} could help prune the graph to only the useful relations. We used simple one-hot edge features throughout, whereas one could use word embeddings \cite{mikolov2013efficient,sa2018representation} to seed the knowledge graph with semantic information.% We could also test on previously published environments such as BoxWorld \cite{Zambaldi2018}, if code becomes available. 

The field has long debated the importance of reasoning with symbols and its compatibility with gradient based learning. Our architecture provides one framework to bridge these seemingly disparate approaches \cite{Garnelo2019}.

%In conclusion, here we show that grounding knowledge graph edges in reinforcement learning tasks can help with fast generalization to tasks involving novel objects.

\bibliography{main}
\bibliographystyle{unsrt}

\appendix

% to compile a preprint version, e.g., for submission to arXiv, add add the
% [preprint] option:
%     \usepackage[preprint]{neurips_2019}

% to compile a camera-ready version, add the [final] option, e.g.:
%     \usepackage[final]{neurips_2019}

% to avoid loading the natbib package, add option nonatbib:
%     \usepackage[nonatbib]{neurips_2019}

\section{Supplemental Methods}

\subsection{Sokoban}

The environment consists of a $10 \times 10$ grid, where the agent is rewarded for pushing balls into their matching buckets. Lower case alphanumeric characters refer to balls, and upper case as buckets. The agent is identified as the \texttt{A} symbol, and the walls with \texttt{+}. For each variation, we generated 100 training mazes, and 20 testing mazes, randomly varying the location of the agent, ball(s), and bucket(s) in each maze. The agent received a reward of $3.0$ for a successful pairing, and a penalty of $-0.1$ for each time step taken.

\subsection{Knowledge graph construction}

The Sokoban games had the edges similar to those shown in Figure 4 of the main text, with an edge feature of `1' from the agent to all balls to encode a 'pushes' relationship; edge feature of `2' between all rewarded ball-bucket pairs; and an edge feature of `0' between the agent and impassable objects: the bucket(s) and the wall symbol. 

In Pacman, we add an 'impassable' relation from all the agents (player, ghost, and scared ghost) to the wall. We also add distinct edges from the player to all edible entities and agents (coin, capsule, scared ghost, ghost).

\subsection{Baseline Algorithms}
We used DQN \cite{Mnih2015}, Prioritized Experience Replay (PER) \cite{schaul2015prioritized}, and A2C \cite{mnih2016asynchronous} as the baseline RL algorithms. To keep the comparison fair, each baseline also received symbolic input. In the Sokoban experiments, we used a convolutional network consisting of  $Conv(3\times3, 64)\rightarrow Conv(3\times3, 64)\rightarrow Dense(64) \rightarrow Dense(4)$. This model is equivalent to the PKGNet architecture with the connections from the knowledge graph removed. We performed an architecture search and did not find a model that outperformed it.

The best agent in Pacman had a deeper and wider convolutional network with four $Conv(3\times3)$ layers with $(64, 128, 128, 64)$ filters, followed by a multilayer perception of $Dense(100) \rightarrow Dense(50) \rightarrow Dense(4)$. 

In both models, after the convolutional layers, we computed a per-channel mean over the 2D map and passed the resulting vector into the multilayer perceptron (MLP).

We validated our implementation of the algorithm by comparing our performance on the Cartpole and Pong environments with those in Coach \cite{caspi_itai_2017_1134899} and Ptan \cite{Lapan2018}. Software was implemented in Pytorch \cite{pytorch} and is attached with the manuscript (see Supplement). OpenAI Gym \cite{Brockman2016} and \textit{pycolab} \cite{pycolab} were used to implement the environments.

\subsection{Hyperparameters}

We ran our experiments using the Adam optimizer with learning rate of $0.0001$ in the Sokoban environments and $0.00025$ in Pacman \cite{Kingma2014}. We used a replay buffer size of 100,000 throughout; at every step, we sampled $32$ transitions from the buffer and trained the agent by minimizing the $L_2$ loss. In the Sokoban environments, we allowed the agent to run for 10,000 steps before commencing training.

\section{Supplemental Figures}
\begin{figure}[tb]
\begin{center}
\centerline{\includegraphics[width=\columnwidth]{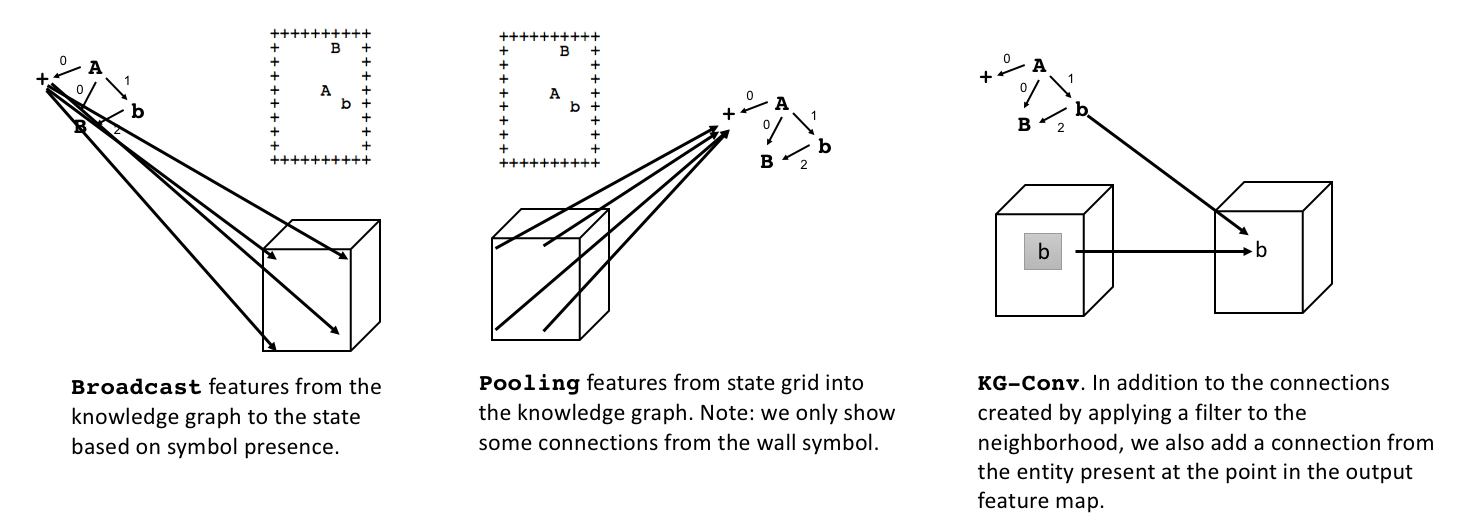}}
\caption{Layer types in our PKGNet architecture, including methods to \texttt{Broadcast} from the knowledge graph $\mathcal{K}$ to the state presentation $\mathcal{S}$, \texttt{Pooling} from $\mathcal{S}\rightarrow\mathcal{K}$, and updating the state by jointly convolving over $(\mathcal{S}, \mathcal{K})$.}
\label{fig:types}
\end{center}
\end{figure}

\begin{figure*}[tb]
\begin{center}
\centerline{\includegraphics[width=\columnwidth]{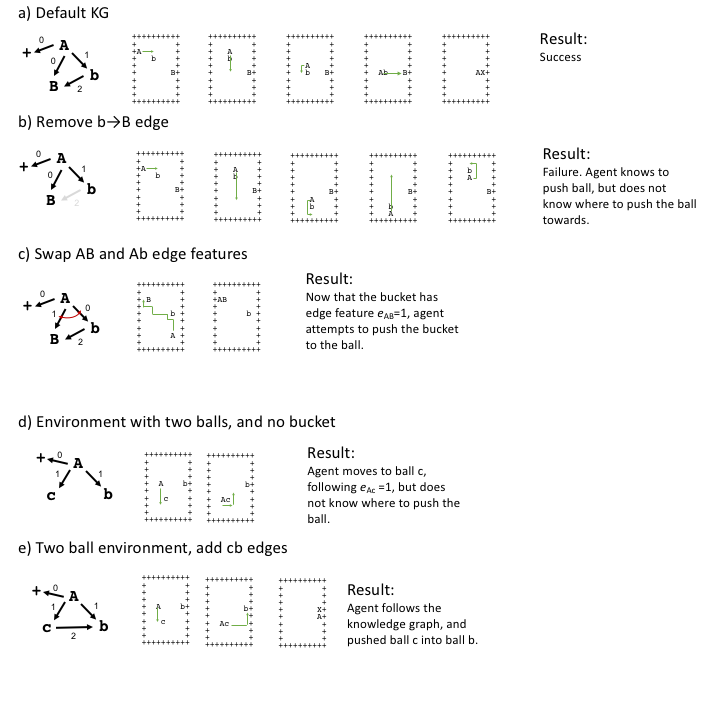}}
\caption{Manipulating agent behavior. We use an already trained agent, and manipulated its behavior at test time by modifying the input knowledge graph. For each manipulation, we show the resulting knowledge graph, the game state, and the resulting agent behavior. These studies show that the agent learned the semantic meaning of edges ('push', 'target') that we intended, and are able to apply those learned relations to different objects. For example, the trained agent can be manipulated to push buckets into balls, or balls into other balls without any additional training.}
\label{fig:manipulate}
\end{center}
\end{figure*}

\begin{figure*}[tb]
\begin{center}
\centerline{\includegraphics[width=\columnwidth]{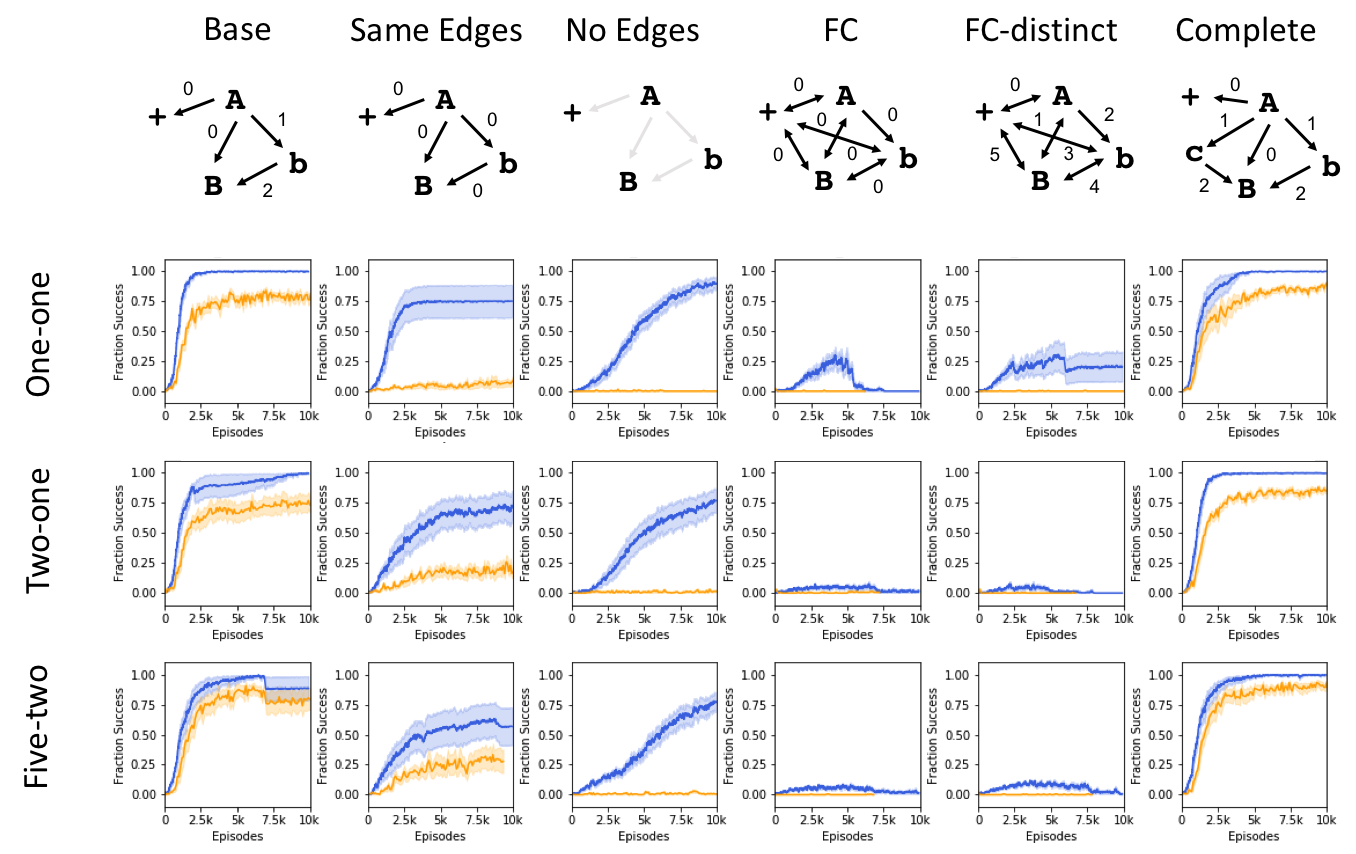}}
\caption{Model performance of PKG-DQN when trained on various knowledge graph types, in the one-one, two-one, and five-two environments. Tested types are described in the paper.}
\label{fig:all}
\end{center}
\end{figure*}

\begin{figure*}[tb]
\begin{center}
\centerline{\includegraphics[width=0.7\columnwidth]{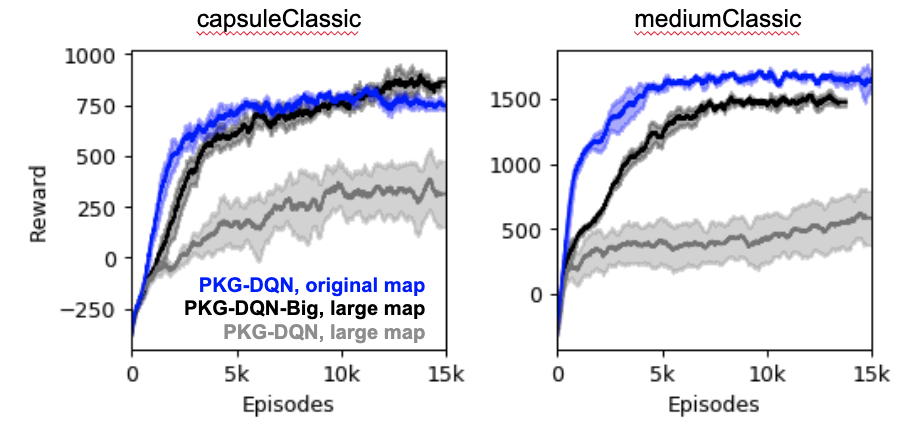}}
\caption{Model performance of PKG-DQN and a larger PKG-DQN. We compare model performance on the original map and a 2x magnified map.}
\label{fig:largepacman}
\end{center}
\end{figure*}

We provide several additional figures that were not included in the main paper:
\begin{itemize}
    \item Figure \ref{fig:types}: Graphical depictions of the three contributed methods (\texttt{Broadcast}, \texttt{Pooling}, and \texttt{KG-Conv} for transferring information between the knowledge graph and the scene representation.
    \item Figure \ref{fig:manipulate}: In addition to Pacman, we also ran experiments with Sokoban where we took an agent trained on the knowledge graph, and observed its behavior when the input knowledge graph was altered. We were able to manipulate the agent behavior, and confirm that the learned edge semantics match the game structure and can be applied to novel objects. Just by changing the knowledge graph at test time, the agent can be manipulated to push buckets into balls, or push balls into other balls.
    \item Figure \ref{fig:all}: A more complete version of Figure 3 from the paper, with additional environments (one-one, two-one) trained with different knowledge graph variants.
    \item Figure \ref{fig:largepacman}: We also tested our PKG-DQN model on a modified version of the Pacman environment that contains extended objects such as those found in natural images. The modified version was created by magnifying each observation by a factor of 2. We were able to match the performance of PKG-DQN on the original map by increasing the size of the model.
\end{itemize}

\section{Model components}

In this section, we provide more details on the model components, as shown in Figure \ref{fig:types}, and described in the main text. We duplicate some of the text from the main paper here for readability.

\paragraph{\bf Broadcast.} We define the function $\texttt{Broadcast}:\mathcal{K}\rightarrow\mathcal{S}$. For each entity $i$ in the knowledge graph, we copy its graph representation $v_i$ to each occurrence of $i$ in the game map. This is used to initialize the state representation $\mathcal{S}$ such that we are using a common embedding to refer to entities in both $\mathcal{K}$ and $\mathcal{S}$. Formally, each location $(i,j)$ in the state is computed as

\begin{equation}
    \mathcal{S}_{i,j} = \sum_{v \in \mathcal{V}} \delta_v(i,j) v
\end{equation}

\noindent where $\delta_v(i,j)=1$ if the entity corresponding to $v$ is present at location $(i,j)$ and zero otherwise. Thus, symbols in the game map not present in the knowledge graph are initialized with a zero vector.

\paragraph{\bf Pooling.} The reverse of \texttt{Broadcast}, this operation is used to update the entity representations in the knowledge graph. In $\texttt{Pooling}:\mathcal{S}\rightarrow\mathcal{K}$, we update the graph's representation $v$ by averaging the features in $\mathcal{S}$ over all instances of entity corresponding to $v$ in the state:

\begin{equation}
    v_i = \frac{1}{N_v}\sum_{(i,j) \in \mathcal{S}} W\delta_v(i,j)\mathcal{S}_{ij}
\end{equation}

\noindent where $N_v=\sum_\mathcal{S} \delta_v(i,j)$ is the number of instances of $v$ in the state. Since $\mathcal{S}$ and $\mathcal{V}$ may have different number of features, we used the weight matrix $W$ to project from the state vectors to the dimensionality of the vertex features in the graph.

\paragraph{\bf KG-Conv.} To update the state representation $\mathcal{S}$, we augment a regular convolution layer with the knowledge graph. In addition to applying convolutional filters to the neighborhood of a location, we also add the node representation $v_i$ of the entity $i$ at that location, passed through linear layer to $v_i$ to match the number of filters in the convolution. Formally, we can describe this operation as:

\begin{equation}
    \text{Conv}_{3\times 3 \times d}(\mathcal{S}) + \text{Conv}_{1\times 1 \times d}(\mathtt{Broadcast}(\mathcal{K}))
\end{equation}

\section{Extended Future Directions}

\subsection{Scenes} The use of scene graphs could provide a framework to handle partial observability by building out portions of the environment as they are explored and storing them in the scene graph. As models that can extract objects from frames improve \cite{Chen2016, Cheung2014, Higgins2016, Ionescu2018, Kingma2013, Locatello2018}, connecting the outputs of these models as inputs to the models developed here could provide a mechanism to go directly from pixels to actions.

\subsection{Interpretability} 

The knowledge graph provides an interpretable way to instruct the Deep RL system the rules of the game. While not explored here these rules could include the model of the environment facilitating use of PKG-DQN in model-based RL. Future work could explore whether the structure of the knowledge graph combined with the interpretability of the nodes and edges could serve as a mechanism to overcome catastrophic forgetting. For example, new entities and relationships could be incrementally added to the knowledge graph encoded in a way that is compatible with existing relationships and with potentially minimal disruption to existing entities and their relationships. A limitation is that even though the knowledge graph itself is interpretable, once the messages from the knowledge graph are combined with messages in the scene graph we sacrifice interpretability in favor of the learning power of gradient based Deep Learning. 

\subsection{Knowledge graph} 

While we are hand coding the knowledge graph in this study, future work could learn the knowledge graph directly from a set of environments, or via information extraction approaches on text corpora, or learn graph attention models over existing large knowledge graphs \cite{Angeli2015, Beetz2015, Bollacker08, Lenat1986, Liu2004, Saxena2014, Suchanek2007}. Knowledge graphs could also be generalized beyond the $\langle subject, relation, object \rangle $ triplet structure to incorporate prior or instructional information in the form of computational graphs. 

\subsection{Environments} While we limited our analysis here to relatively small environments to test the fundamental aspects of our approach, scaling to larger environments is another obvious direction. Environments such as OpenAI Retro \cite{Nichol2018} or CoinRun \cite{cobbe2018quantifying} have helped spark an interest in the problem of generalization in Deep RL. However, the lack of readily available ground truth and inability to programmatically generate levels hinders a rigorous development of algorithmic approaches to solve this problem using Retro. We believe that further development of benchmarks for generalization in Deep RL \cite{Packer2018} that enable programmatic game creation and make ground truth accessible will help the field.

% \section{Instructions for Code}
% We have attached a full copy of our experiment code to our submission. The installation instructions and commands to run the experiments can be found in the \texttt{README} file. We have also included a docker template that builds the repository along with its dependencies, as well as a set of commands to train the model and reproduce the figures in the main paper.

\clearpage

\end{document}